\documentclass{article}
\usepackage[utf8]{inputenc}
\usepackage{main}
\usepackage{microtype}
\usepackage{graphicx}
\usepackage{times}
\usepackage{latexsym}
\usepackage{amsmath}
\usepackage{float}
\usepackage{footnote}
\usepackage{enumitem}
\usepackage{bm}
\usepackage{arydshln}
\usepackage{booktabs}
\usepackage{multicol}
\usepackage{multirow}
\usepackage{color}
\usepackage{xcolor}     
\usepackage{colortbl}
\usepackage{bbding}
\usepackage{makecell}
\usepackage{mathtools}
\usepackage{imakeidx}
\usepackage{longtable}
\usepackage{wrapfig}
\makeindex
\usepackage{arydshln}
\usepackage{lipsum}
\usepackage{natbib}
\usepackage[toc]{multitoc}
\usepackage[edges]{forest}
\usepackage[normalem]{ulem}
\definecolor{mydarkblue}{rgb}{0,0.08,0.45}
\usepackage[colorlinks=true,linkcolor=mydarkblue,citecolor=mydarkblue,filecolor=mydarkblue,urlcolor=mydarkblue]{hyperref}
\usepackage{caption}
\usepackage{CJKutf8}
\usepackage{awesomebox} 
\usepackage{bbding}
\usepackage[most]{tcolorbox}

\usepackage{subcaption}
\usepackage{tabularx}

\usepackage[tikz]{bclogo}
\usepackage[framemethod=tikz]{mdframed}
\definecolor{bgblue}{RGB}{245,243,253}
\definecolor{ttblue}{RGB}{91,194,224}

\mdfdefinestyle{mystyle}{%
  rightline=true,
  innerleftmargin=10,
  innerrightmargin=10,
  outerlinewidth=3pt,
  topline=false,
  rightline=true,
  bottomline=false,
  skipabove=\topsep,
  skipbelow=\topsep
}

\newtcolorbox{myboxi}[1][]{
  breakable,
  title=#1,
  colback=red!5,
  colbacktitle=red!5,
  coltitle=black,
  fonttitle=\bfseries,
  bottomrule=0pt,
  toprule=0pt,
  leftrule=2pt,
  rightrule=2pt,
  titlerule=0pt,
  arc=0pt,
  outer arc=0pt,
  colframe=red,
}

\newtcolorbox{myboxnote}[1][]{
  breakable,
  title=#1,
  colback=orange!0,
  colbacktitle=orange!0,
  coltitle=black,
  fonttitle=\bfseries,
  bottomrule=0pt,
  toprule=0pt,
  leftrule=2pt,
  rightrule=2pt,
  titlerule=0pt,
  arc=0pt,
  outer arc=0pt,
  colframe=orange,
}

\newtcolorbox{myboxii}[1][]{
  breakable,
  freelance,
  title=#1,
  colback=white,
  colbacktitle=white,
  coltitle=black,
  fonttitle=\bfseries,
  bottomrule=0pt,
  boxrule=0pt,
  colframe=white,
  overlay unbroken and first={
  \draw[red!75!black,line width=3pt]
    ([xshift=5pt]frame.north west) -- 
    (frame.north west) -- 
    (frame.south west);
  \draw[red!75!black,line width=3pt]
    ([xshift=-5pt]frame.north east) -- 
    (frame.north east) -- 
    (frame.south east);
  },
  overlay unbroken app={
  \draw[red!75!black,line width=3pt,line cap=rect]
    (frame.south west) -- 
    ([xshift=5pt]frame.south west);
  \draw[red!75!black,line width=3pt,line cap=rect]
    (frame.south east) -- 
    ([xshift=-5pt]frame.south east);
  },
  overlay middle and last={
  \draw[red!75!black,line width=3pt]
    (frame.north west) -- 
    (frame.south west);
  \draw[red!75!black,line width=3pt]
    (frame.north east) -- 
    (frame.south east);
  },
  overlay last app={
  \draw[red!75!black,line width=3pt,line cap=rect]
    (frame.south west) --
    ([xshift=5pt]frame.south west);
  \draw[red!75!black,line width=3pt,line cap=rect]
    (frame.south east) --
    ([xshift=-5pt]frame.south east);
  },
}

\usepackage{fancyhdr} 
\usepackage{blindtext} 

\newcommand{\widtha}{0.325\textwidth}
\newcolumntype{C}{>{\centering\arraybackslash}X}

\DeclareCaptionFont{black}{\color{black}}

\definecolor{myblue}{rgb}{0.9, 0.1, 0.94}
\definecolor{mygreen}{rgb}{0.64, 0.56, 0.88}
\definecolor{myyellow}{rgb}{0.68, 0.6, 0.1}
\definecolor{fancygreen}{rgb}{0.33, 0.68, 0.20}
\definecolor{salmon}{rgb}{0.94, 0.52, 0.49}
\definecolor{tablegreen}{rgb}{0.82, 0.94, 0.75}
\definecolor{tableblue}{rgb}{0.81, 0.90, 0.94}
\definecolor{tablered}{rgb}{0.97, 0.85, 0.85}
\definecolor{tableorange}{rgb}{0.96, 0.85, 0.81}

\newenvironment{itemize*}%
 {\leftmargini=10pt\begin{itemize}%
  \setlength{\itemsep}{0pt}%
  \setlength{\parskip}{0pt}%
  }%
 {\end{itemize}}
\newenvironment{enumerate*}%
 {\begin{enumerate}%
  \setlength{\itemsep}{0pt}%
  \setlength{\parskip}{0pt}}%
 {\end{enumerate}}

\usepackage{xcolor}
\usepackage{listings}

\newcommand\JSONnumbervaluestyle{\color{blue}}
\newcommand\JSONstringvaluestyle{\color{red}}

\newif\ifcolonfoundonthisline

\makeatletter

\lstdefinestyle{json}
{
  showstringspaces    = false,
  keywords            = {false,true},
  alsoletter          = 0123456789.,
  morestring          = [s]{"}{"},
  stringstyle         = \ifcolonfoundonthisline\JSONstringvaluestyle\fi,
  MoreSelectCharTable =%
    \lst@DefSaveDef{`:}\colon@json{\processColon@json},
  basicstyle          = \ttfamily,
  keywordstyle        = \ttfamily\bfseries,
}

\newcommand\processColon@json{%
  \colon@json%
  \ifnum\lst@mode=\lst@Pmode%
    \global\colonfoundonthislinetrue%
  \fi
}

\lst@AddToHook{Output}{%
  \ifcolonfoundonthisline%
    \ifnum\lst@mode=\lst@Pmode%
      \def\lst@thestyle{\JSONnumbervaluestyle}%
    \fi
  \fi
  \lsthk@DetectKeywords%
}

\lst@AddToHook{EOL}%
  {\global\colonfoundonthislinefalse}

\makeatother

\usepackage{etoolbox}
\usepackage{natbib}
\usepackage{url}
\newcounter{bibcount}
\makeatletter
\patchcmd{\@lbibitem}{\item[}{\item[\hfil\stepcounter{bibcount}{[\thebibcount]}}{}{}
\renewcommand\NAT@bibsetup%
  [1]{\setlength{\leftmargin}{\bibhang}\setlength{\itemindent}{-\parindent}%
      \setlength{\itemsep}{\bibsep}\setlength{\parsep}{\z@}}
\makeatother

\begin{document}

\title{\textsc{Anole}: Open Autoregressive Multimodal Models for  Image-Text Generation (without Diffusion)}

\title{\raisebox{-.25ex}{\includegraphics[height=2ex]{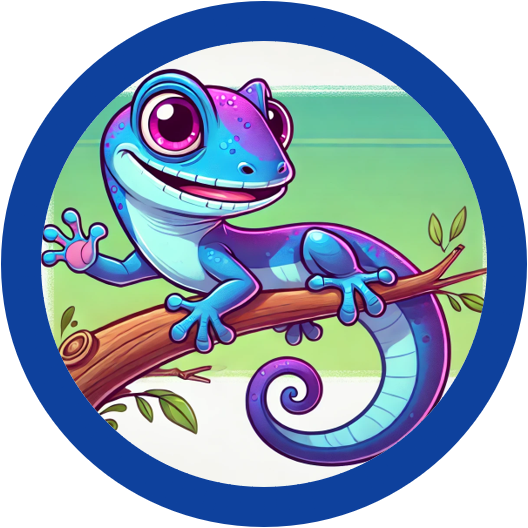}}\textsc{Anole}: An Open, Autoregressive, Native Large Multimodal Models for  Interleaved Image-Text Generation}

\newcommand*\samethanks[1][\value{footnote}]{\footnotemark[#1]}

\newcommand{\modelname}{\textsc{Anole}\xspace}

\author{Ethan Chern\thanks{~~Co-first authors.}\space\space \space\space \space\space\space
Jiadi Su\samethanks\space\space \space\space \space\space\space
Yan Ma\samethanks\space\space \space\space \space\space\space
\textbf{Pengfei Liu}\thanks{~~Corresponding author}\\ \\
\textbf{Generative AI Research Lab (GAIR)}
}

\maketitle

\thispagestyle{fancy}
\fancyhead{}
\lhead{\includegraphics[height=0.67cm]{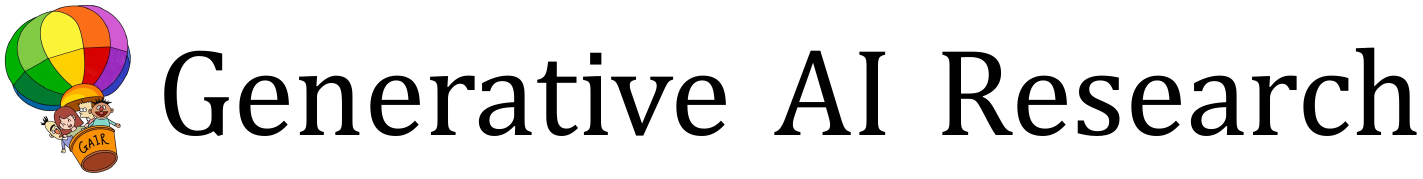}}
\renewcommand{\headrulewidth}{0pt}
\setlength{\headsep}{0mm}

\begin{abstract}
Previous open-source large multimodal models (LMMs) have faced several limitations: (1) they often lack native integration, requiring adapters to align visual representations with pre-trained large language models (LLMs); (2) many are restricted to single-modal generation; (3) while some support multimodal generation, they rely on separate diffusion models for visual modeling and generation. To mitigate these limitations, we present \modelname, an open, autoregressive, native large multimodal model for interleaved image-text generation. We build \modelname from Meta AI's Chameleon, adopting an innovative fine-tuning strategy that is both data-efficient and parameter-efficient. \modelname demonstrates high-quality, coherent multimodal generation capabilities. We have open-sourced our model, training framework, and instruction tuning data.

\raisebox{-0.2\height}{\includegraphics[height=0.40cm]{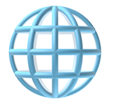}}\textbf{Homepage}: \url{https://gair-nlp.github.io/anole}

\raisebox{-0.12\height}{\hspace{0.03cm}\includegraphics[height=0.34cm]{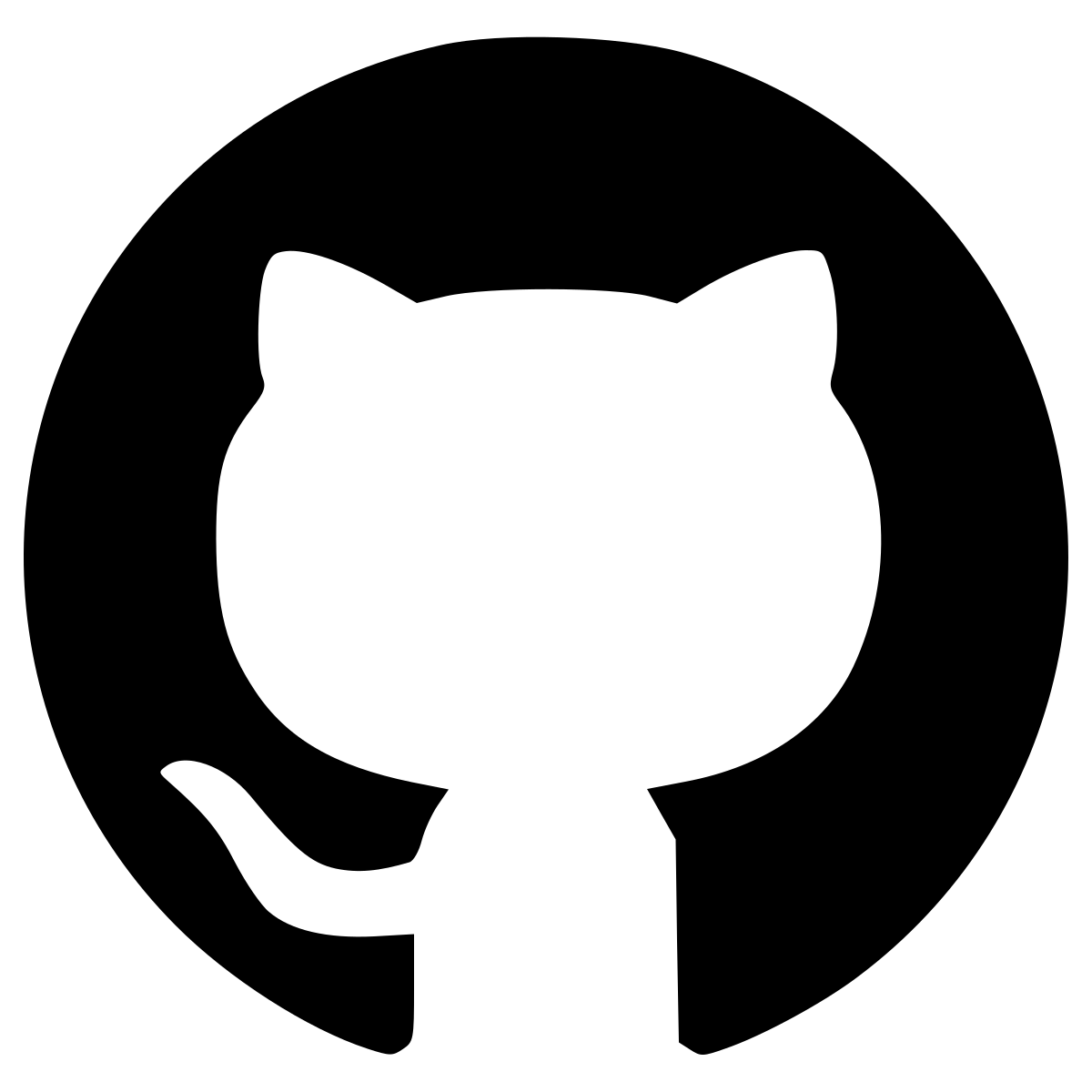}} \textbf{Code}: \url{https://github.com/GAIR-NLP/anole}

\raisebox{-0.2\height}{\includegraphics[height=0.40cm]{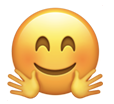}}\textbf{Hugging Face Model}: \url{https://huggingface.co/GAIR/Anole-7b-v0.1}

\end{abstract}

\begin{figure}[ht]
    \centering
    \includegraphics[page=1, width=0.85\textwidth]{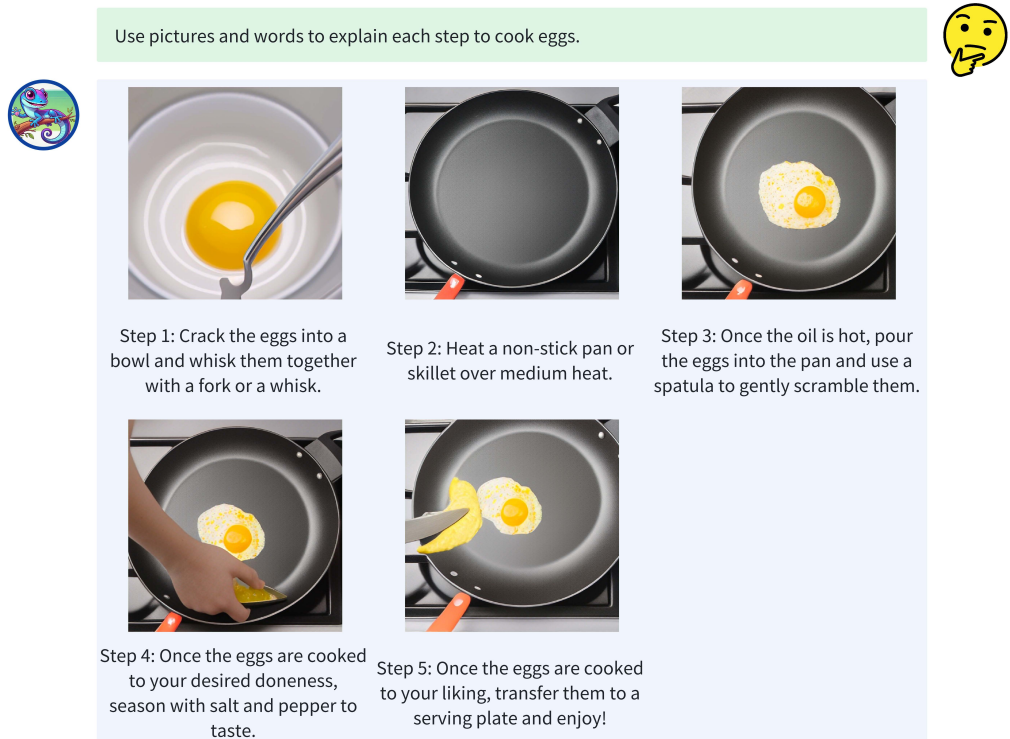}
    \caption{An example of \modelname generating a high-quality and coherent interleaved image-text sequence on how to cook eggs.
    }
    \label{fig:interleaved1}
\end{figure}
 
\newpage

\pagestyle{fancy}
\lhead{\rightmark}
\renewcommand{\headrulewidth}{0.7pt}
\setlength{\headsep}{5mm}



\section{Introduction}
Since the introduction of Meta AI's LLaMA \cite{touvron2023llama} in February 2023, autoregressive open-source large language models (LLMs) such as LLaMA \cite{touvron2023llama}, Alpaca \cite{taori2023stanford}, Vicuna \cite{vicuna2023}, Mistral \cite{jiang2023mistral}, Phi \cite{gunasekar2023textbooks}, Gemma \cite{team2024gemma}, Olmo \cite{groeneveld2024olmo} and LLM360 \cite{liu2023llm360} have democratized and advanced the development of LLMs. Efforts in open-sourcing large multimodal models (LMMs) are also ongoing, though at a slower pace compared to LLMs. Notable open-sourced LMMs include LLaVA \cite{liu2023llava,liu2023improvedllava}, CogVLM \cite{wang2023cogvlm}, DreamLLM \cite{dong2023dreamllm}, Emu2 \cite{sun2024generative} and Cambrian \cite{tong2024cambrian}. However, current open-source LMMs have several significant limitations: (1) Many focus solely on multimodal understanding without multimodal generation \cite{liu2023llava, liu2023improvedllava, wang2023cogvlm}, (2) most are not natively multimodal (i.e., not trained on multimodal data from the pretraining stage) and rely on pretrained LLMs as their backbone \cite{liu2023llava, liu2023improvedllava, wang2023cogvlm, dong2023dreamllm, sun2024generative}, and (3) those with vision generation capabilities require an additional diffusion model for vision modeling and generation \cite{dong2023dreamllm, sun2024generative}. This reliance on extra mechanisms can introduce complexity and inefficiency in both training and inference time. 

Given the limitations in current open-source LMMs, the AI community eagerly anticipates the emergence of truly open, autoregressive, native LMMs with multimodal generation capabilities. The goal is to be able to develop LMMs in the same way we do with LLMs. To address this gap, we introduce \modelname. As shown in Fig.~\ref{fig:interleaved1}, \modelname can generate a high-quality, coherent recipe for cooking eggs in just a few seconds.

\modelname is built on top of Chameleon~\citep{team2024chameleon} by Meta AI. Chameleon represents a significant advancement in multimodal AI, showcasing the potential of early-fusion, token-based autoregressive approaches for multimodal modeling. According to their paper, Chameleon has demonstrated impressive capabilities in understanding and generating interleaved sequences of images and text, pushing the boundaries of what is possible in multimodal AI. The latest open-source release of Chameleon has shown strong performance in text understanding, text generation, and multimodal comprehension. However, the current open-source version of Chameleon \textbf{does not} support image generation or multimodal generation. We build \modelname to facilitate Chameleon's capabilities in vision and multimodal generation without compromising its strengths in text generation and multimodal comprehension. \modelname addresses the limitations of the current open-source release of Chameleon, making the full potential of multimodal generation accessible to the broader research community. The key contributions of our work are fourfold:

\begin{enumerate}
    \item Full Open-Source Implementation: \modelname has facilitated the vision and multimodal generation capabilities from Chameleon through an innovative fine-tuning approach, unlocking the model's most crucial technological aspects. This comprehensive open-source release allows researchers and developers to fully utilize and build upon it.
    \item Data and Parameter Efficient Fine-Tuning: Our method fine-tunes fewer than 40M parameters, requiring only about 6,000 samples to effectively facilitate vision and multimodal generation capabilities. This demonstrates a highly efficient approach to facilitate complex functionality in LMMs.
    \item Training, Multimodal Inferece, and Qualitative Evaluation: We provide a training and multimodal inference framework for unified tokenizer-based multimodal models. This infrastructure significantly lowers the barrier to entry for developing and experimenting with autoregressive LMMs, making it accessible to a wider range of researchers. Additionally, we conduct qualitative analysis to demonstrate the potential of autoregressive LMMs.
    \item Rich Resources for Accessibility: To further support the adoption and advancement of autoregressive LMMs, we offer an extensive collection of data resources and detailed tutorials. These materials are designed to facilitate easier onboarding and experimentation for researchers at various levels of expertise.
\end{enumerate}

By addressing these critical aspects, \modelname represents a significant step forward in democratizing access to advanced multimodal AI technologies. Our work not only builds upon the foundations laid by the original Chameleon model but also paves the way for more inclusive and collaborative research in the field of multimodal AI.
 
Moreover, \modelname sparks a series of important and intriguing research questions for the community to explore. For example:
\begin{itemize}
    \item Investigating the performance limits of vision generation using unified tokenizer-based multimodal models, in comparison to established methods like diffusion models.
    \item Developing efficient techniques for interleaved image-text decoding, which are essential for real-world applications such as textbook and comic generation.
    \item Exploring optimal fine-tuning methodologies for these complex pretrained LMMs.
    \item Addressing critical issues, including ensuring the safety and ethical use of generated images.
\end{itemize}

\section{Related Works}

\begin{table}[htbp]
\centering
\footnotesize
\begin{tabular}{lcccc}
\toprule
\textbf{Models}  &\textbf{w/o Diffusion?} &\textbf{Native?}
&\textbf{Token-based?} &\textbf{Interleaved (open-source)?} \\
\midrule
MM-Interleaved \cite{tian2024mm}  &\XSolidBrush   &\XSolidBrush   &\XSolidBrush &\Checkmark \\
Emu2 \cite{sun2024generative}            &\XSolidBrush   &\XSolidBrush   &\XSolidBrush &\XSolidBrush \\
DreamLLM \cite{dong2023dreamllm}        &\XSolidBrush   &\XSolidBrush   &\XSolidBrush &\XSolidBrush \\
AnyGPT  \cite{zhan2024anygpt}         &\XSolidBrush   &\XSolidBrush   &\Checkmark &\XSolidBrush \\
LWM \cite{liu2024world}             &\Checkmark     &\XSolidBrush   &\Checkmark &\XSolidBrush \\
Chameleon \cite{team2024chameleon}      &\Checkmark     &\Checkmark     &\Checkmark &\XSolidBrush  \\
\midrule
\modelname (Ours)          &\Checkmark     &\Checkmark     &\Checkmark &\Checkmark  \\
\bottomrule
\end{tabular}
\caption{\textbf{Comparison with related works that allows text and (or) vision generation.} \textbf{w/o Diffusion} indicates whether the model relies on a diffusion model to generate visual contexts. \textbf{Native} specifies if the model is a native LMM. \textbf{Token-based} denotes whether the model utilizes token-based modeling approach. \textbf{Interleaved (open-source)} represents whether the open-source released version of the model supports multimodal generation.}
\label{tab:datasets}
\end{table}

LMMs have experienced significant advancements, often leveraging pretrained LLMs as their foundation backbone. One common approach involves combining a CLIP-pretrained vision encoder with diffusion models as the decoder, and a pretrained LLM (e.g., Vicuna) as the backbone \cite{dong2023dreamllm, tian2024mm, sun2024generative}. This approach achieves impressive results in both image understanding and generation by leveraging the robust representations provided by CLIP and diffusion models. However, incorporating CLIP and diffusion models increases the overall complexity of the model architecture, resulting in additional training overhead and reduced inference efficiency.

In contrast, purely token-based methods exclude diffusion models and CLIP, instead using token-based representations for multimodal understanding and generation. This approach traces back to BEiT \cite{bao2021beit}, with OpenAI's DALL-E \cite{ramesh2021zero} exemplifying text-to-image generation based on similar principles. These methods rely heavily on vector quantization (VQ) models \cite{van2017neural, esser2021taming}, which combine ResNet-based encoders and decoders with a discrete codebook. During image encoding, the VQ model transforms the image from pixel space to latent space representations, then maps these representations to codebook IDs using a nearest-neighbor search. These IDs serve as input tokens for a Transformer model, which models conditional probabilities and predicts sequences. The VQ decoder then reconstructs images from the generated sequences. This autoregressive, discrete image encoding and decoding approach has been validated in multiple studies for producing high-quality images \cite{zhu2024scaling, yu2024image}, effectively modeling inter-image dependencies \cite{bai2023sequential}, and enhancing image consistency \cite{pan2024synthesizing}. LWM \cite{liu2024world} and Chameleon \cite{team2024chameleon} extend this concept to image-text multimodal tasks, using streamlined architectures to handle tasks involving both images and text. Compared to other methods, unified token-based modeling significantly reduces model complexity, facilitating seamless inference and the generation of interleaved image-text sequences without additional components.

\modelname, building on the foundation of Chameleon, facilitates Chameleon's image and multimodal generation capabilities with efficient fine-tuning. \modelname retains the inherent advantages of Chameleon's architecture while producing high-quality images and maintaining coherent image-text sequences. Tab.~\ref{tab:datasets} highlights \modelname's characteristics as an autoregressive, diffusion-free, native token-based model, emphasizing its capability to support multimodal generation, as well as its simplicity and efficiency compared to more complex frameworks.

\section{\modelname}

\begin{figure}[ht]
    \centering
    \includegraphics[width=\textwidth]{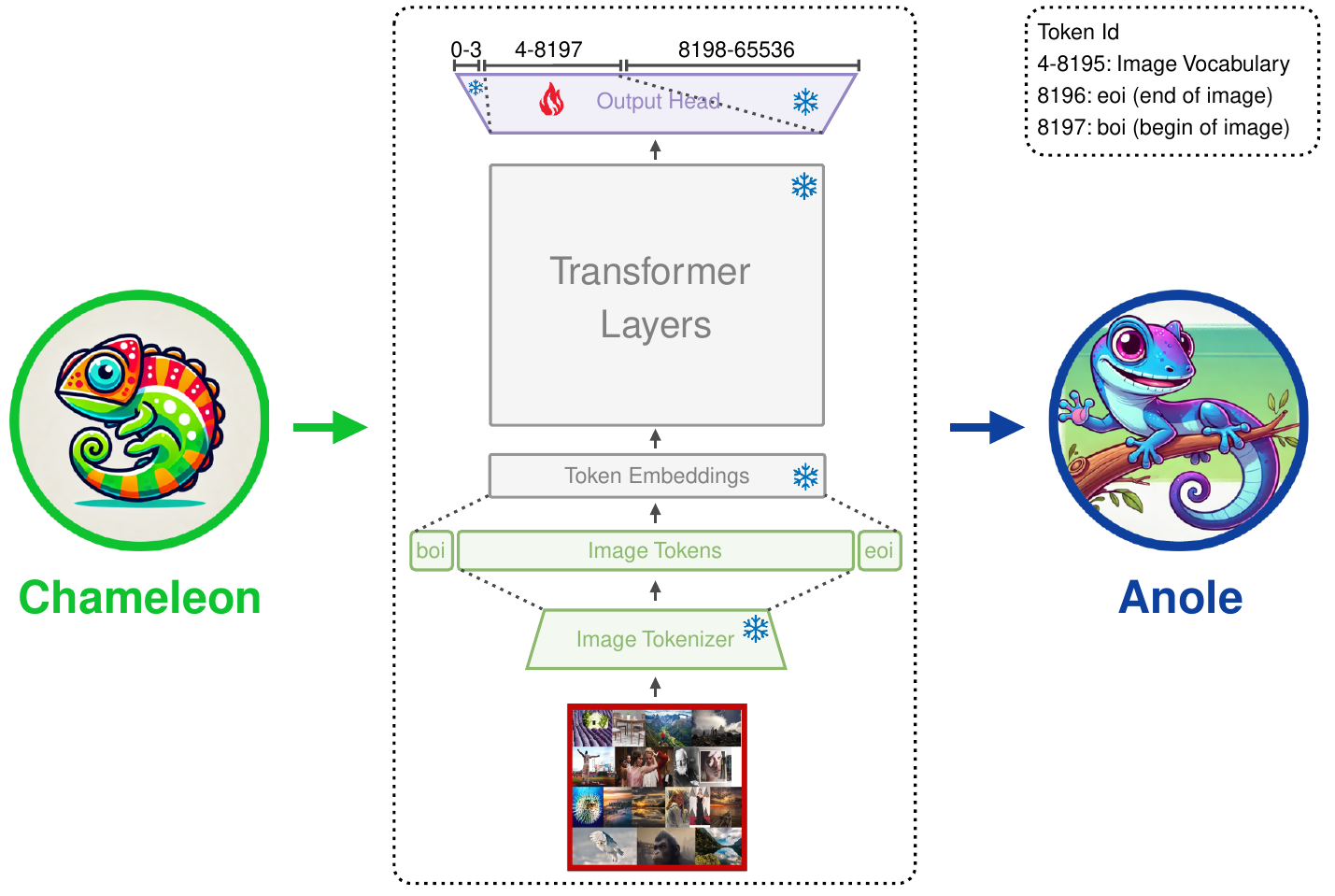}
    \caption{Facilitating Chameleon's image generation capabilities via innovative fine-tuning.}
    \label{fig:anole_main}
\end{figure}

\subsection{Overview}
\modelname adopts the same approach and architecture as Chameleon, utilizing an early-fusion, token-based, autoregressive approach to model multimodal sequences (text and images) without the use of diffusion models, relying solely on transformers. Token-based approaches \cite{team2024chameleon, lu2022unified, yu2023scaling, liu2024world} achieve modality fusion at the input-token level. Firstly, modality-specific tokenizers tokenize samples from each modality. Then, these token sequences are concatenated to form a single multimodal token sequence, which is subsequently fed into an autoregressive transformer for modeling.

\subsection{Facilitating the image generation and multimodal generation capabilities from Chameleon}
Fig.~\ref{fig:anole_main} shows the innovative fine-tuning process of how \modelname is fine-tuned from Chameleon. Based on available information and our testings, the latest release of Chameleon have demonstrated strong performance in text understanding, text generation, and multimodal understanding. \modelname, build on top of Chameleon, aiming to facilitate the image generation and multimodal generation capabilities from Chameleon. Chameleon’s pre-training data natively includes both text and image modalities, theoretically equipping it with image generation capabilities. Our goal is to facilitate this ability without compromising its text understanding, generation, and multimodal comprehension. To achieve this, we froze most of Chameleon’s parameters and fine-tuned only the logits corresponding to image token ids in transformer’s output head layer.

Following the principle of ``less is more'' \cite{zhou2024lima}, the current version of \modelname, \modelname-7b-v0.1, was developed using a small amount of image data (5,859 images from LAION-5B art \cite{schuhmann2022laion}) and was fine-tuned on just a few parameters (less than 40M) in a short time (around 30 minutes on 8 A100 GPUs). Despite these limitations, \modelname-7b-v0.1 expresses impressive image (Tab.~\ref{tab:text2image} and Tab.~\ref{tab:other1}) and multimodal generation capabilities (Fig.~\ref{fig:interleaved1},  Fig.~\ref{fig:interleaved2} and Fig.~\ref{fig:overall}).

\section{Evaluation} 
We conduct qualitative analysis on \modelname's capabilities on image generation and interleaved image-text generation.

\subsection{Image Generation}
Tab.~\ref{tab:text2image} demonstrates the text-to-image capabilities of \modelname. We highlight the following points:
(1) The images generated by \modelname are of high quality and closely adhere to the given instructions. For instance, \modelname accurately captures the essence of "A steaming cup of coffee next to a fresh croissant on a cozy cafe table," showcasing the steam, coffee, and croissant elements precisely.
(2) \modelname demonstrates remarkable versatility in generating diverse types of images. It can create realistic depictions, as seen in the coffee and ice cream images, as well as imaginative scenes, like the dinosaur strolling in Times Square. This variety highlights \modelname's ability to blend realism with creativity seamlessly.

\begin{table}[htbp]
\centering
\begin{tabularx}{0.99\textwidth}{*{3}{C}}
\toprule
        A serene lakeside view at sunrise with mist rising from the water, surrounded by dense pine forests and mountains in the background. & A bustling downtown street in Tokyo at night, with neon  signs, crowded sidewalks, and tall skyscrapers. &  A colorful ice cream sundae topped with sprinkles, whipped cream, and a cherry. \\
\midrule
\includegraphics[width=\widtha]{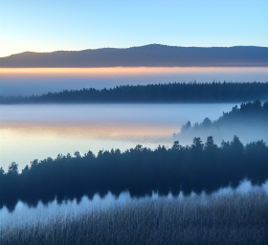} & \includegraphics[width=\widtha]{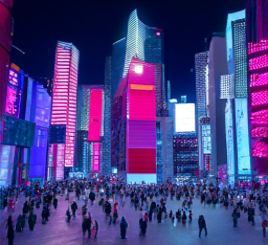} & \includegraphics[width=\widtha]{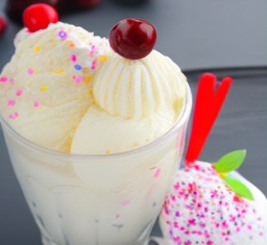} \\   
\includegraphics[width=\widtha]{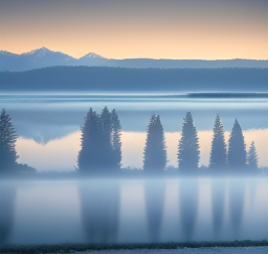} & \includegraphics[width=\widtha]{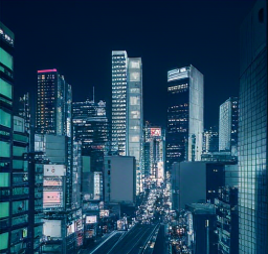} & \includegraphics[width=\widtha]{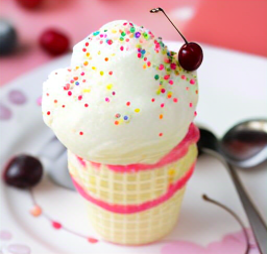} \\
\midrule        
        A steaming cup of coffee next to a fresh croissant on a cozy café table. & An impressionist painting of a café scene, with loose brushstrokes and lively colors. & Under the bright sun, dinosaurs strolling in Times Square, with people taking photos and traffic passing by.  \\
\midrule   
\includegraphics[width=\widtha]{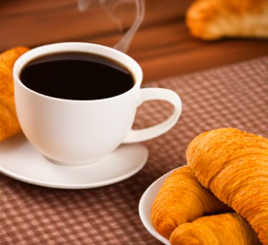} & \includegraphics[width=\widtha]{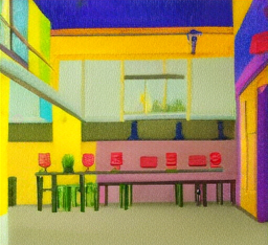} & \includegraphics[width=\widtha]{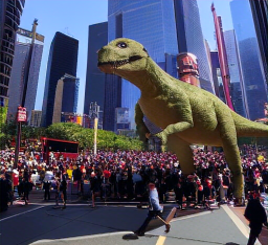} \\
\includegraphics[width=\widtha]{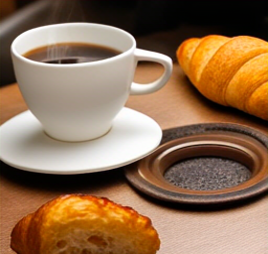} & \includegraphics[width=\widtha]{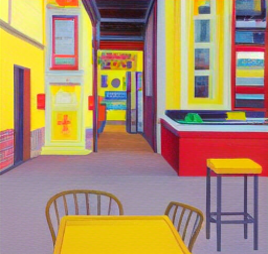} & \includegraphics[width=\widtha]{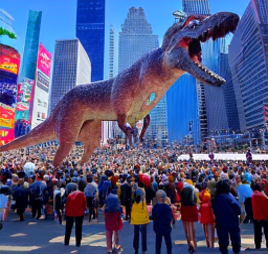} \\
\bottomrule
\end{tabularx}
\vspace{-3px}
\caption{Text-to-image examples generated by \modelname.}
\label{tab:text2image}
\end{table}

\begin{table}[htbp]
\centering
\begin{tabularx}{0.99\textwidth}{*{3}{C}}
\toprule
        A piece of paper with word ``Anole'' written on it. & A piece of paper with word like ``Anole'' written on it, and a drawing of an Anole. &  An image depicting three cubes stacked on a table. Each cube has a unique color and a letter on it. \\
\midrule
\includegraphics[width=\widtha]{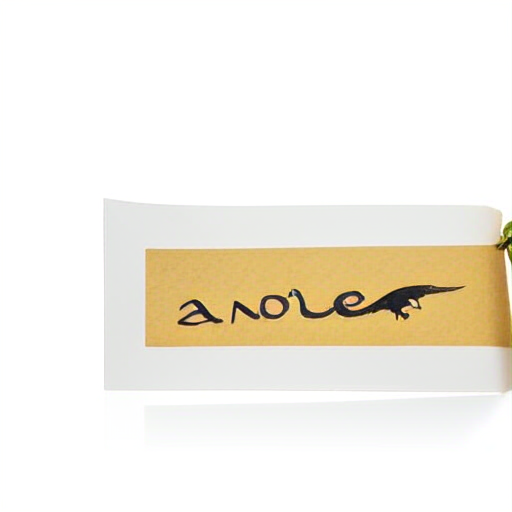} & \includegraphics[width=\widtha]{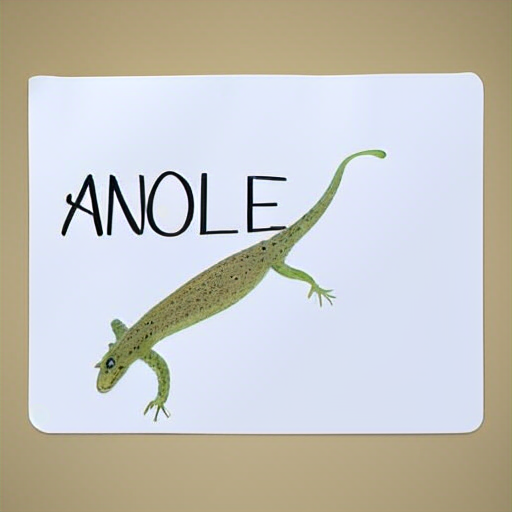} & \includegraphics[width=\widtha]{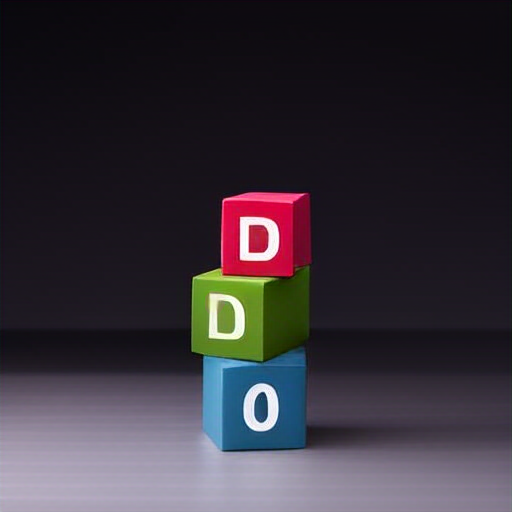} \\
\midrule        
        A vibrant coral reef teeming with colorful fish, sea turtles gliding through the water, and sunlight filtering down from the surface. & A tropical beach with crystal clear water, white sand, and palm trees swaying in the breeze. & A quiet European village with cobblestone streets and colorful houses, under a clear blue sky.  \\
\midrule   
\includegraphics[width=\widtha]{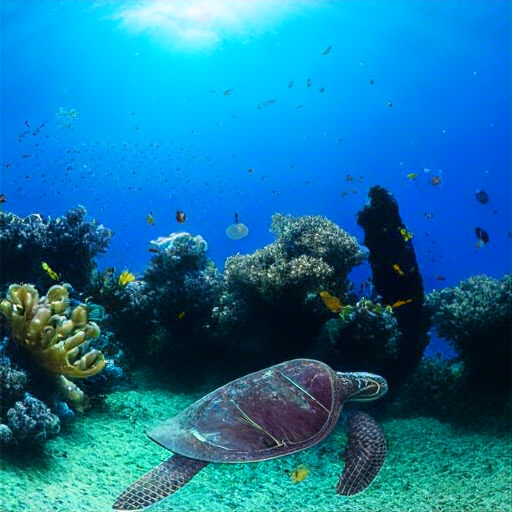} & \includegraphics[width=\widtha]{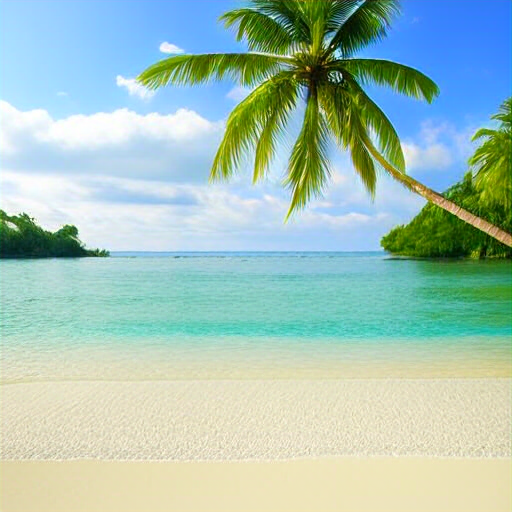} & \includegraphics[width=\widtha]{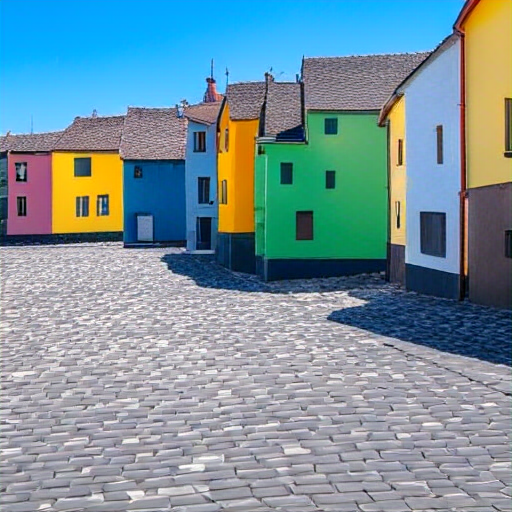} \\
\midrule        
        A gothic cathedral with intricate stone carvings and stained glass windows. & A delicious slice of pizza with melting cheese and pepperoni, on a rustic wooden table. & A colorful graffiti mural on an urban wall, with vibrant characters and abstract elements.  \\
\midrule   
\includegraphics[width=\widtha]{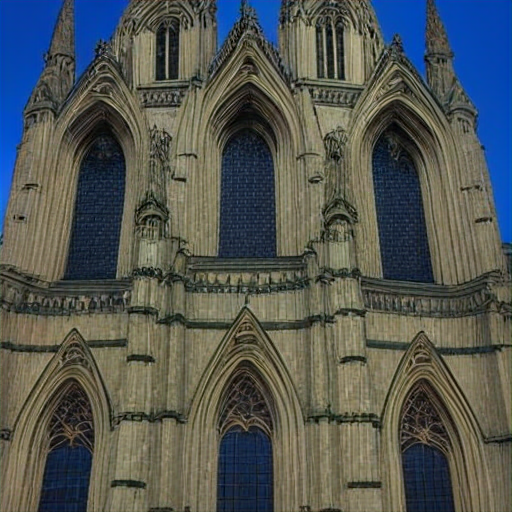} & \includegraphics[width=\widtha]{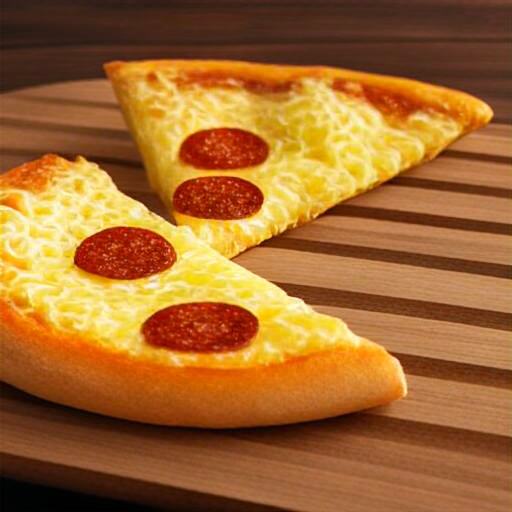} & \includegraphics[width=\widtha]{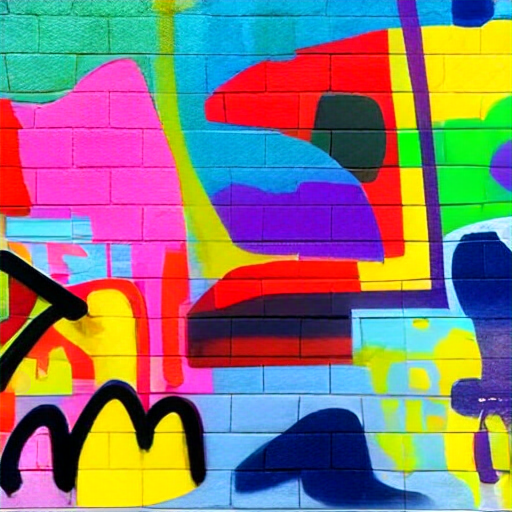} \\
\bottomrule
\end{tabularx}
\vspace{-3px}
\caption{More text-to-image examples generated by \modelname.}
\label{tab:other1}
\end{table}

\subsection{Interleaved Image-Text Generation}
\begin{figure}[ht]
    \centering
    \includegraphics[page=2, width=0.85\textwidth]{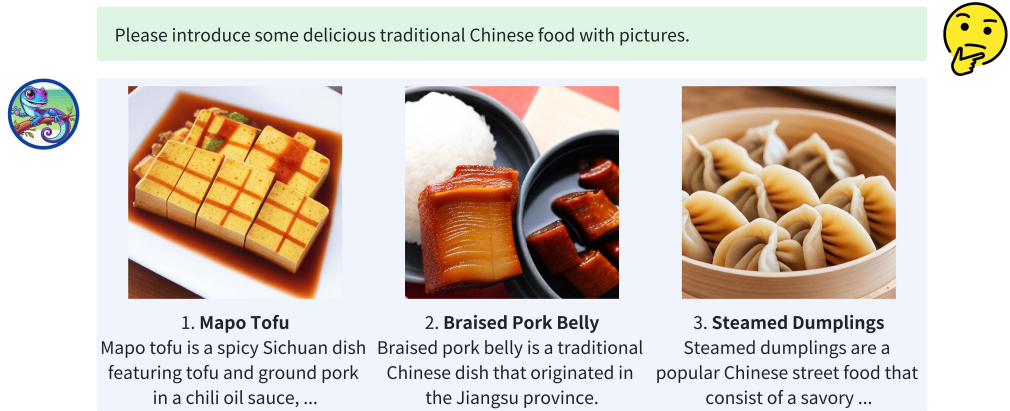}
    \caption{An example of \modelname generating an interleaved image-text sequence on how to cook traditional Chinese cuisines.}
    \label{fig:interleaved2}
\end{figure}

\begin{figure}[htbp]
    \centering
    \begin{subfigure}[b]{0.9\textwidth}
        \centering
        \begin{subfigure}[b]{\textwidth}
            \centering
            \includegraphics[width=\textwidth]{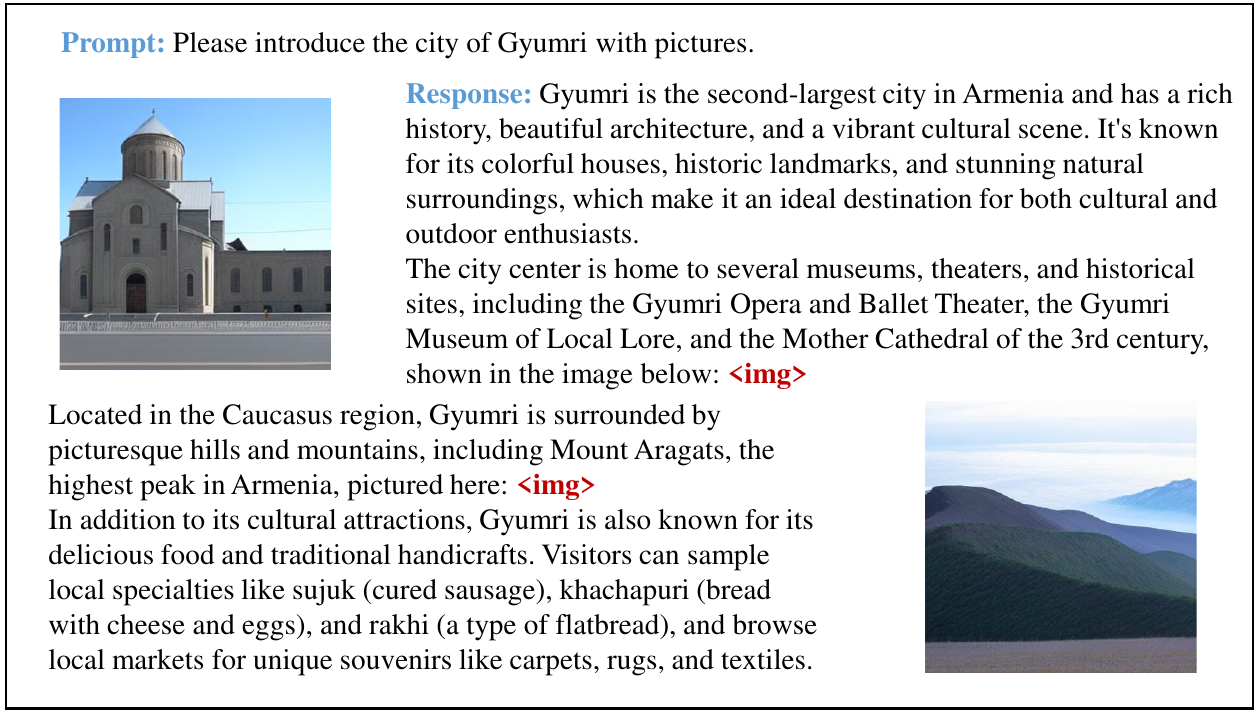}
            \label{fig:other2}
        \end{subfigure}
        \vfill
        \begin{subfigure}[b]{\textwidth}
            \centering
            \includegraphics[width=\textwidth]{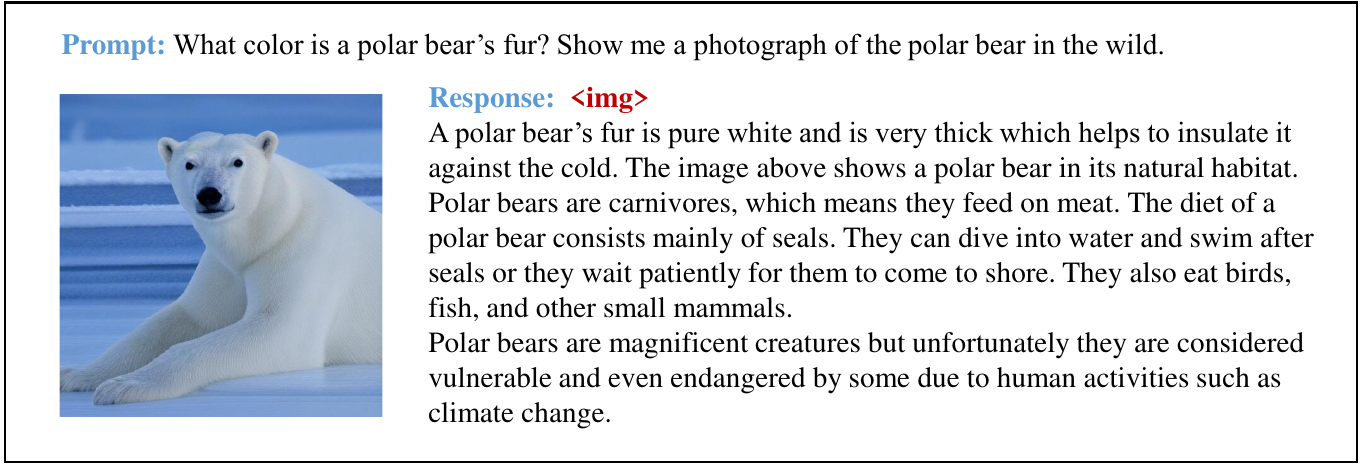}
            \label{fig:other3}
        \end{subfigure}
    \end{subfigure}
    \caption{Examples of \modelname generating interleaved text-and-image sequences on geographical and biological topics.}
    \label{fig:overall}
\end{figure}

Fig.~\ref{fig:interleaved1},  Fig.~\ref{fig:interleaved2} and Fig.~\ref{fig:overall} demonstrate the multimodal generation capabilities of \modelname. We emphasize that (1) The generated text is well-organized and provides comprehensive details, which is essential for accurate interleaved image-text generation. This is evident in Fig.~\ref{fig:overall}, where the detailed introduction of Gyumri is accompanied by relevant images, effectively capturing the city's architectural design and cultural highlights. (2) \modelname demonstrates seamless integration between images and text, ensuring that the visual and textual elements complement each other perfectly. Fig.~\ref{fig:interleaved2} highlights this capability by introducing traditional Chinese dishes with corresponding images, making the information both engaging and informative.

\section{Conclusion \& Future Directions}
We introduce \modelname, an open, autoregressive, native LMM for interleaved image-text generation that demonstrates advanced multimodal generation abilities. \modelname facilitates image and multimodal generation capabilities from Chameleon by fine-tuning on just 6,000 samples with 40M parameters. We are committed to continually upgrading \modelname to enhance its capabilities. Our future directions include (1) enhancing \modelname's precise instruction-following capability, (2) extending its context length, (3) improving its multimodal understanding capabilities, and (4) applying \modelname to downstream tasks requiring multimodal generation abilities.

\section{Limitations \& Disclaimer}
\modelname is intended for research use only. Our model weights follow the same license as Chameleon. The fine-tuning images we used are from LAION-5B art, and thus follow the same license as LAION. \modelname is still under development and has many limitations that need to be addressed. Importantly, we have not aligned the image generation capabilities of the \modelname to ensure safety and harmlessness. Therefore, we encourage users to interact with \modelname with caution and report any concerning behaviors to help improve the model's safety and ethical considerations.

\section{Acknowledgements}
We sincerely thank the Meta Chameleon Team for open-sourcing Chameleon. Our base model and the majority of our inference code are based on it. We also greatly appreciate all the contributors who have participated in pull request \#31534 submitted to the transformers repository. This PR has been crucial for the development of our training code.

\newpage
\bibliography{main}
\bibliographystyle{acl_natbib}

\end{document}